\documentclass[10pt,twocolumn,letterpaper]{article}

\usepackage{cvpr}              % To produce the CAMERA-READY version
\usepackage{graphicx}
\usepackage{amsmath}
\usepackage{amssymb}
\usepackage{booktabs}
\usepackage{glossaries}

\usepackage[accsupp]{axessibility}  % Improves PDF readability for those with disabilities.
\newcommand\scalemath[2]{\scalebox{#1}{\mbox{\ensuremath{\displaystyle #2}}}}

\usepackage{textcomp}
\usepackage{gensymb}
\usepackage{bm}
\usepackage{siunitx}
\usepackage{xcolor}
\usepackage{float}
\usepackage{mathtools}
\newacronym{rl}{RL}{Reinforcement Learning}
\newacronym{ppo}{PPO}{Proximal Policy Optimization}
\newacronym{ddppo}{DD-PPO}{Decentralized Distributed Proximal Policy Optimization}
\newacronym{rnn}{RNN}{Recurrent Neural Network}
\newacronym{lstm}{LSTM}{Long Short-Term Memory}
\newacronym{mdp}{MDP}{Markov Decision Process}
\newacronym{pomdp}{POMDP}{Partially Observable Markov Decision Process}

\newacronym{vit}{ViT}{Vision Transformer}
\newacronym{ma}{MHA}{Multi-head Attention}
\newacronym{msa}{MHSA}{Multi-head Self-Attention}
\newacronym{sa}{SA}{Self-Attention}
\newacronym{cls}{$[CLS]$}{Classification Token}
\newacronym{act}{$[ACT]$}{Action Token}
\newacronym{mae}{MAE}{Masked Autoencoder}
\newacronym{mmae}{MultiMAE}{Multi-modal Multi-task Masked Autoencoder}
\newacronym{dino}{DINO}{Self-DIstillation and NO Labels}
\newacronym{vot}{VOT}{Visual Odometry Transformer}

\newacronym{nlp}{NLP}{Natural Language Processing}
\newacronym{bert}{BERT}{Bidirectional Encoder Representations from Transformers}
\newacronym{gpt}{GPT}{Generative Pre-trained Transformer}
\newacronym{nsp}{NSP}{Next Sentence Prediction}
\newacronym{mlm}{MLM}{Masked Language Model}
\newacronym{seq2seq}{Seq2Seq}{Sequence to Sequence}

\newacronym{cv}{CV}{Computer Vision}
\newacronym{cnn}{ConvNet}{Convolution Neural Network}
\newacronym{resnet}{ResNet}{Residual Network}
\newacronym{mlp}{MLP}{Multi-layer Perceptron}
\newacronym{ssl}{SSL}{Self-supervised Learning}
\newacronym{cbam}{CBAM}{Convolutional Block Attention Module}
\newacronym{bn}{BN}{Batch Normalization}
\newacronym{cbn}{CBN}{Conditional Batch Normalization}

\newacronym{ai}{AI}{Artificial Intelligence}
\newacronym{habitat}{Habitat}{AI Habitat}
\newacronym{slam}{SLAM}{Simultaneous Localization and Mapping}
\newacronym{vo}{VO}{Visual Odometry}
\newacronym{gdsp}{GDSP}{Geodesic Distance along the Shortest Path}
\newacronym{spl}{SPL}{Success weighted by (normalized inverse) Path Length}
\newacronym{softspl}{SSPL}{Soft Success Path Length}
\newacronym{imu}{IMU}{Inertial Measurement Unit}

\newacronym{gelu}{GELU}{Gaussian Error Linear Unit}
\newacronym{relu}{ReLU}{Rectified Linear Unit}
\newacronym{timm}{timm}{PyTorch Image Models}
\newacronym{sgd}{SGD}{Stochastic Gradient Descent}
\newacronym{in}{ImageNet}{ImageNet}
\newacronym{in1k}{IN-1k}{ImageNet-1k}
\newacronym{in21k}{IN-21k}{ImageNet-21k}
\newacronym{knn}{KNN}{K-nearest Neighbors}

\newacronym{semseg}{\texttt{SemSeg}}{\texttt{Semantic Segmenation}}
\newacronym{icp}{ICP}{Iterative Closest Point}

\newcommand{\fwd}[0]{{\small\texttt{fwd}}\xspace}
\newcommand{\turnl}[0]{{\small\texttt{left}}\xspace}
\newcommand{\turnr}[0]{{\small\texttt{right}}\xspace}
\newcommand{\stp}[0]{\texttt{stop}\xspace}

\newcommand{\gps}[0]{\texttt{GPS+Compass}\xspace}
\newcommand{\rgbd}{\texttt{RGB-D}\xspace}
\newcommand{\rgb}{\texttt{RGB}\xspace}
\newcommand{\depth}{\texttt{Depth}\xspace}

\newcommand{\blind}{\texttt{Blind}\xspace}
\newcommand{\lidar}{\texttt{LIDAR}\xspace}

\usepackage{pifont}% Zapf-Dingbats Symbole
\newcommand*{\cmark}{\ding{52}}

\usepackage[pagebackref,breaklinks,colorlinks]{hyperref}
\hypersetup{ colorlinks=true, linkcolor=blue, filecolor=magenta, urlcolor=cyan, }

\usepackage[capitalize]{cleveref}
\crefname{section}{Sec.}{Secs.}
\Crefname{section}{Section}{Sections}
\Crefname{table}{Table}{Tables}
\crefname{table}{Tab.}{Tabs.}

\def\cvprPaperID{7066} % *** Enter the CVPR Paper ID here
\def\confName{CVPR}
\def\confYear{2023}

\begin{document}

\title{Modality-invariant Visual Odometry for Embodied Vision \vspace{-4mm}}

%\author{Marius Memmel\thanks{University of Washington, work done on exchange at EPFL}\\
%{\tt\small memmelma@cs.washington.edu}
%\and
%Roman Bachmann\thanks{Swiss Federal Institute of Technology Lausanne (EPFL)}\\
%{\tt\small roman.bachmann@epfl.ch}
%\and
%Amir Zamir\footnotemark[2]\\
%{\tt\small amir.zamir@epfl.ch}
%}
\author{Marius Memmel\textsuperscript{1}\thanks{Work done on exchange at EPFL} \quad\quad Roman Bachmann\textsuperscript{2} \quad\quad Amir Zamir\textsuperscript{2} \\ 
\textsuperscript{1}University of Washington \quad\quad \textsuperscript{2}Swiss Federal Institute of Technology (EPFL)\vspace{3mm}\\
\url{https://vo-transformer.github.io}
}
\maketitle

\begin{abstract}
\vspace{-2.0mm}
Effectively localizing an agent in a realistic, noisy setting is crucial for many embodied vision tasks. Visual Odometry (VO) is a practical substitute for unreliable GPS and compass sensors, especially in indoor environments. While SLAM-based methods show a solid performance without large data requirements, they are less flexible and robust \wrt to noise and changes in the sensor suite compared to learning-based approaches. Recent deep VO models, however, limit themselves to a fixed set of input modalities, \eg, RGB and depth, while training on millions of samples. When sensors fail, sensor suites change, or modalities are intentionally looped out due to available resources, \eg, power consumption, the models fail catastrophically. Furthermore, training these models from scratch is even more expensive without simulator access or suitable existing models that can be fine-tuned. While such scenarios get mostly ignored in simulation, they commonly hinder a model's reusability in real-world applications. We propose a Transformer-based modality-invariant VO approach that can deal with diverse or changing sensor suites of navigation agents. Our model outperforms previous methods while training on only a fraction of the data. We hope this method opens the door to a broader range of real-world applications that can benefit from flexible and learned VO models.
\end{abstract}
\vspace{-2.em}
\section{Introduction}
Artificial intelligence has found its way into many commercial products that provide helpful digital services.
To increase its impact beyond the digital world, personal robotics and embodied AI aims to put intelligent programs into bodies that can move in the real world or interact with it~\cite{Duan2022}.
One of the most fundamental skills embodied agents must learn is to effectively traverse the environment around them, allowing them to move past stationary manipulation tasks and provide services in multiple locations instead~\cite{Savva2019}. The ability of an agent to locate itself in an environment is vital to navigating it successfully~\cite{Datta2020,Zhao2021}.
A common setup is to equip an agent with an \rgbd (\rgb and \depth) camera and a \gps sensor and teach it to navigate to goals in unseen environments~\cite{Anderson2018}. With extended data access through simulators~\cite{Savva2019,Szot2021,Xia2020,Savva2017,Kolve2017}, photo-realistic scans of 3D environments~\cite{Kolve2017,Chang2017,Straub2019,Xia2018,Wu2018}, and large-scale parallel training, recent approaches reach almost perfect navigation results in indoor environments~\cite{Wijmans2020}.
However, these agents fail catastrophically in more realistic settings with noisy, partially unavailable, or failing \rgbd sensor readings, noisy actuation, or no access to \gps~\cite{HabitatLeaderboard2020,Zhao2021}.

\begin{figure}[t!]
    \centering
    \includegraphics[width=\linewidth]{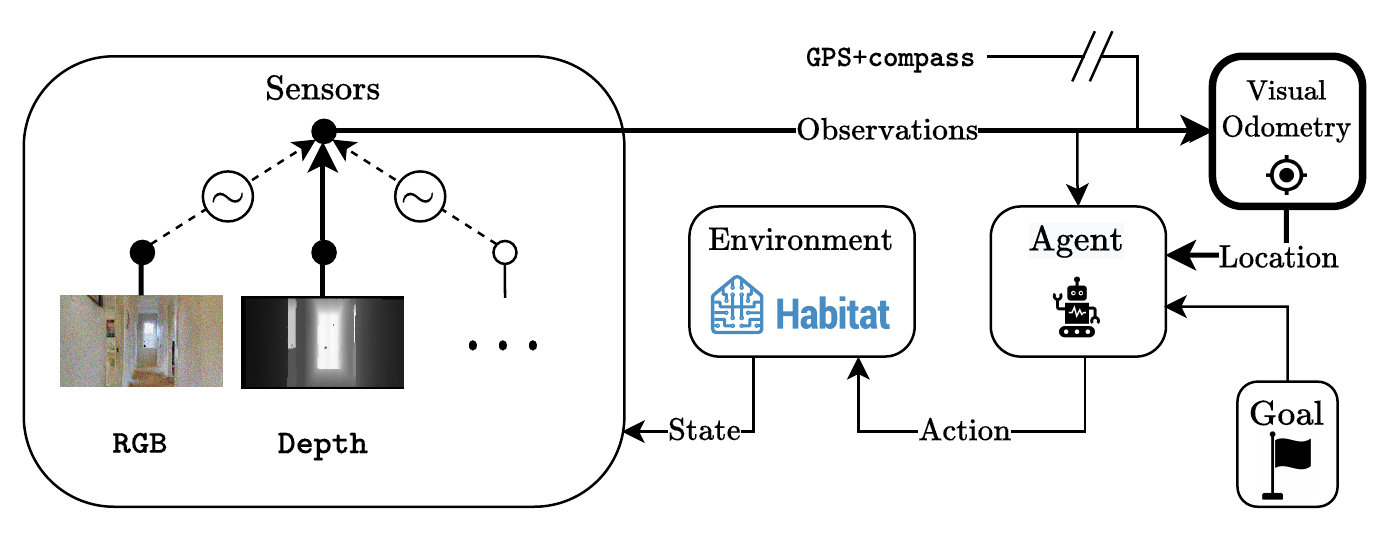}
    \caption{An agent is tasked to navigate to a goal location using \rgbd sensors. Because \gps are not available, the location is inferred from visual observations only. Nevertheless, sensors can malfunction, or availability can change during test-time (indicated by $\sim$), resulting in catastrophic failure of the localization. We train our model to react to such scenarios by randomly dropping input modalities. Furthermore, our method can be extended to learn from multiple arbitrary input modalities, \eg, surface normals, point clouds, or internal measurements.}
    \label{fig:overview}
    \vspace{-1.em}
\end{figure}

% visual odometry and failing sensors
\gls{vo} is one way to close this performance gap and localize the agent from only \rgbd observations~\cite{Anderson2018}, and deploying such a model has been shown to be especially beneficial when observations are noisy~\cite{Datta2020,Zhao2021}.
However, those methods are not robust to any sensory changes at the test-time, such as a sensor failing, underperforming, or being intentionally looped out.
In practical applications~\cite{Srinivasa2019}, low-cost hardware can also experience serious bandwidth limitations, causing \rgb (3 channels) and \depth (1 channel) to be transferred at different rates. Furthermore, mobile edge devices must balance battery usage by switching between passive (\eg, \rgb) and active (\eg, \lidar) sensors depending on the specific episode.
% In practical applications~\cite{Srinivasa2019}, \eg, using on-board computers like the \textit{Nvidia Jetson Nano} in conjunction with the \textit{Intel RealSense Depth Camera D435}, \rgb is only available for 85\% of \depth observations.
Attempting to solve this asymmetry by keeping separate models in memory, relying on active sensors, or using only the highest rate modality is simply infeasible for high-speed and real-world systems.
Finally, a changing sensor suite represents an extreme case of sensor failure where access to a modality is lost during test-time.
These points demonstrate the usefulness of a certain level of modality invariance in a VO framework.
Those scenarios decrease the robustness of SLAM-based approaches~\cite{Mishkin2019} and limit the transferability of models trained on \rgbd to systems with only a subset or different sensors.

We introduce \textit{``optional" modalities} as an umbrella term to describe settings where input modalities may be of limited availability at test-time. \Cref{fig:overview} visualizes a typical indoor navigation pipeline, but introduces uncertainty about modality availability (\ie at test-time, only a subset of all modalities might be available). While previous approaches completely neglect such scenarios, we argue that explicitly accounting for ``optional" modalities already \emph{during training} of \gls{vo} models allows for better reusability on platforms with different sensor suites and trading-off costly or unreliable sensors during test-time.
Recent methods~\cite{Datta2020,Zhao2021} use \gls{cnn} architectures that assume a constant channel size of the input, which makes it hard to deal with multiple "optional" modalities.
In contrast, Transformers~\cite{Vaswani2017} are much more amenable to variable-sized inputs, facilitating the training of models that can optionally accept one or multiple modalities~\cite{Bachmann2022}.

% data requirements
%Training a Transformer from scratch requires massive amounts of data due to the lack of inductive bias in contrast to \glspl{cnn}~\cite{Dosovitskiy2020,Zhai2021}. 
%Compared to \glspl{cnn}~\cite{Dosovitskiy2020,Zhai2021}, Transformers have fewer inductive biases, which makes using large amounts of data necessary for training from scratch.
Transformers are known to require large amounts of data for training from scratch.
Our model's data requirements are significantly reduced by incorporating various biases: 
We utilize multi-modal pre-training~\cite{Likhosherstov2021,Girdhar2022,Bachmann2022}, which not only provides better initializations but also improves performance when only a subset of modalities are accessible during test-time~\cite{Bachmann2022}. 
%By introducing different biases into our model, we drastically reduce its data requirements. First, we build on multi-modal representation learning~\cite{Likhosherstov2021,Girdhar2022,Bachmann2022}. Besides providing advanced model initializations, these methods improve performance when only a subset of the modalities is available at test-time~\cite{Bachmann2022}
Additionally, we propose a token-based action prior. The action taken by the agent has shown to be beneficial for learning \gls{vo}~\cite{Zhao2021,Partsey2021} and primes the model towards the task-relevant image regions.

% % multi-modality and 
% Access to multiple modalities was also shown to be beneficial for many downstream tasks~\cite{Sax2020,Likhosherstov2021,Girdhar2022}, particularly for \gls{vo}~\cite{Zhu2022,Valada2018,Radwan2018}. Recent work has investigated learning representations from multiple modalities~\cite{Likhosherstov2021,Girdhar2022,Bachmann2022}. These methods provide advanced model initializations, reduce data requirements, and may also improve performance when only a subset of the modalities is available at test-time~\cite{Bachmann2022}. Additionally, the action taken by the agent represents a powerful prior on the \gls{vo} task~\cite{Zhao2021,Partsey2021}. Conditioning the model on this information supports the learning process and further reduces the data requirements.

We introduce the \gls{vot}, a novel modality-agnostic framework for \gls{vo} based on the Transformer architecture. Multi-modal pre-training and an action prior drastically reduce the data required to train the architecture. Furthermore, we propose explicit modality-invariance training. By dropping modalities during training, a single \gls{vot} matches the performance of separate uni-modal approaches. This allows for traversing different sensors during test-time and maintaining performance in the absence of some training modalities. 

We evaluate our method on point-goal navigation in the \textit{Habitat Challenge 2021}~\cite{Habitatchallenge2020sim2real} and show that \gls{vot} outperforms previous methods~\cite{Partsey2021} with training on only 5\% of the data. Beyond this simple demonstration, we stress that our framework is modality-agnostic and not limited to \rgbd input or discrete action spaces and can be adapted to various modalities, \eg, point clouds, surface normals, gyroscopes, accelerators, compass, etc. 
To the best of our knowledge, \gls{vot} is the first widely applicable modality-invariant Transformer-based \gls{vo} approach and opens up exciting new applications of deep \gls{vo} in both simulated and real-world applications.
We make our code available at \href{https://github.com/memmelma/VO-Transformer}{github.com/memmelma/VO-Transformer}.

\begin{figure*}[t!]
    \centering
    \includegraphics[width=0.7\textwidth]{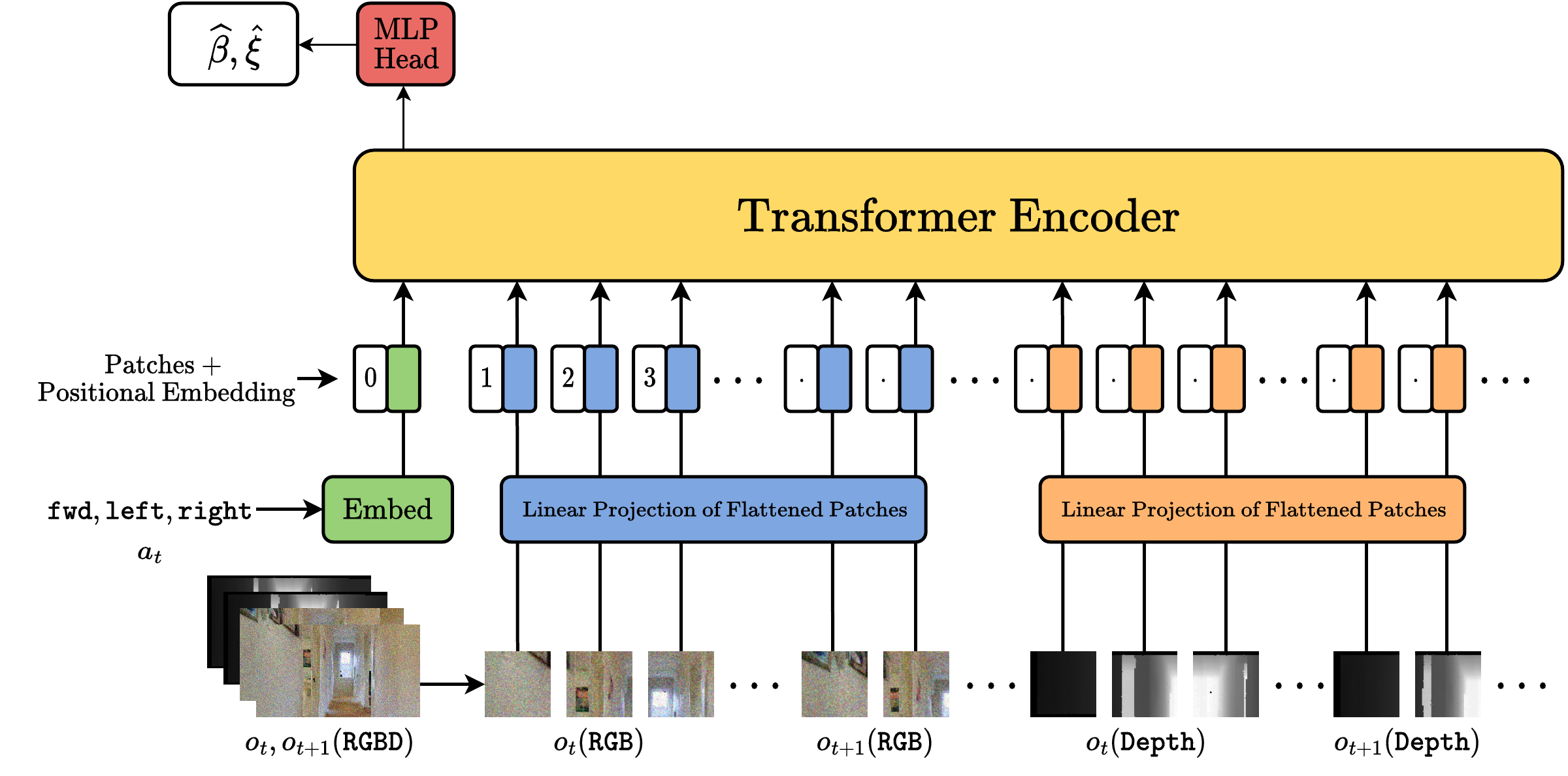}
    \caption{The \glsentrylong{vot} architecture for \rgbd input. Image patches are turned into tokens through modality-specific linear projections
    {\color[HTML]{7EA6E0}{\rule{0.7em}{0.7em}}} {\color[HTML]{FFB570}{\rule{0.7em}{0.7em}}} before a fixed positional embedding is added to them. We pass an action token that embeds the action {\color[HTML]{97D077}{\rule{0.7em}{0.7em}}} taken by the agent as we find it acts as a strong prior on the \gls{vo} problem. An \glsentryshort{mlp}-head {\color[HTML]{EA6B66}{\rule{0.7em}{0.7em}}} then estimates the \gls{vo} parameters $\widehat{\beta},\widehat{\bm{\xi}}$, i.e., translation and rotation of the agent, from the output token. By randomly dropping either \rgb or \depth during training, the Transformer backbone {\color[HTML]{FFD966}{\rule{0.7em}{0.7em}}} becomes modality-agnostic, allowing it to deal with a subset of these input modalities during test-time without losing performance. When more modalities are available during training, other modality-specific linear projections can be added to process the additional information.}
    \label{fig:architecture}
    \vspace{-1.em}
\end{figure*}

\section{Related Work}
\noindent\textbf{\glsentryshort{slam}- vs Learning-based Navigation:}
\gls{slam} approaches decompose the navigation task into the components of mapping, localization, planning, and control~\cite{Thrun2002}.
These methods rely on explicit visual feature extraction and, therefore, fail in realistic settings with noisy observations~\cite{Zhao2021}, while learning-based methods are more robust to noise, ambiguous observations, and limited sensor suites~\cite{Kojima2019,Mishkin2019}.
However, learning-based methods require an order of magnitude more data, \eg, available through simulation~\cite{Savva2019}.
To deal with the large data requirements, \gls{slam}- and learning-based methods can be combined~\cite{Chaplot2020a,Chaplot2020b,Bhatti2016,Zhang2017,Zhan2020,Teed2021,Czarnowski2020}.

\noindent\textbf{Visual Odometry for Realistic Indoor Navigation:}
While most \gls{vo} methods estimate an agent's pose change from more than two frames~\cite{Wang2017,Wang2018} or optical flow~\cite{Zhu2022}, subsequent frames in indoor environments share almost no overlap and contain many occlusions due to the large displacement caused by the discrete action space~\cite{Zhao2021}.
Datta \etal~\cite{Datta2020} propose to estimate the pose change from consecutive frames via a \gls{cnn} architecture and decouple learning the \gls{vo} from the task-specific navigation policy to allow for retraining modules when dynamics change or the actuation experiences noise.
Zhao \etal~\cite{Zhao2021} improve the model's robustness to observation and actuation noise through geometric invariance losses~\cite{Wang2019}, separate models for moving and turning, pre-process observations, and introduce dropout~\cite{Srivastava2014}.
Finally, Partsey \etal~\cite{Partsey2021} explore the need for explicit map building in autonomous indoor navigation. They apply train- and test-time augmentations and concatenate an action embedding similar to Zhao \etal~\cite{Zhao2021} to the extracted visual features.
A trend is to exploit simulators to gather large datasets (1M~\cite{Zhao2021}, 5M~\cite{Partsey2021}). While this is a reasonable progression, it is infeasible to re-train the \gls{vo} model whenever dynamics or sensor configurations change.

\noindent\textbf{Multi-modal Representation Learning:}
The availability of multi-modal or pseudo-labeled~\cite{Bachmann2022} data~\cite{Deng2009, Silberman2012, Zamir2018, Zhou2019, Eftekhar2021,Roberts2021}, \eg, depth, video, and audio, makes it possible to learn feature-rich representations over multiple modalities. Together with Transformer's~\cite{Vaswani2017} ability to process a token sequence of arbitrary length, this leads to general-purpose architectures that can handle various modalities~\cite{Jaegle2021} like video, images, and audio~\cite{Likhosherstov2021} or single-view 3D geometry~\cite{Girdhar2022}.
In particular, \gls{mmae}~\cite{Bachmann2022} is a multi-modal pre-training strategy that performs masked autoencoding~\cite{He2022} with \rgb, \depth, and \gls{semseg}. We show that fine-tuning a pre-trained \gls{mmae} model can significantly increase \gls{vo} performance using only 5\% of the training data amount of previous methods~\cite{Partsey2021}.

\section{Proposed Method}

\subsection{Preliminaries}
In the realistic PointGoal Navigation task~\cite{Anderson2018}, an agent spawns at a random position in an unseen environment and is given a random goal location $\bm{g}_t$ relative to its starting position. At each time step $t$ of an episode, the agent perceives its environment through observations $\bm{o}_t$ and executes an action $a_t$ from a set of discrete actions (move \fwd $0.25m$, turn \turnl and \turnr by $30\degree$). The \stp action indicates the agent's confidence in having reached the goal.
Because the relative goal position $\bm{g}_t$ is defined at the beginning of each episode, it has to be updated throughout the episode as the actions change the agent's position and orientation.
Following \cite{Datta2020,Zhao2021}, we update $\bm{g}_t$ through an estimate of the agent's coordinate transformation.
With access to \gps, computing this transformation is trivial. However, since those sensors are unavailable, we estimate the transformation from the agent's subsequent observations $\bm{o}_t,\bm{o}_{t+1}$ and update the estimated relative goal position $\widehat{\bm{g}}_t$.
When taking an action $a_t$, the agent's coordinate system $C_t$ transforms into $C_{t+1}$. Because the agent can only navigate planarly in the indoor scene, we discard the 3rd dimension for simplicity.
We define the estimated transformation as $\bm{\widehat{H}} \in SE(2)$, with $SE(2)$ being the group of rigid transformations in a 2D plane and parameterize it by the estimated rotation angle $\widehat{\beta} \in \mathbb{R}$ and estimated translation vector $\bm{\widehat{\xi}} \in \mathbb{R}^2$:
% We estimate these \gls{vo} parameters via
\begin{equation}\label{eq:vo_parameters}
    % \begin{split}
        % &
        \scalemath{0.95}{
        \bm{\widehat{H}} =
            \begin{bmatrix}
                \widehat{R} & \bm{\widehat{\xi}}\\ 
                0 & 1
            \end{bmatrix}, \quad
        \widehat{R} =
            \begin{bmatrix*}[r]
                \cos(\widehat{\beta}) & -\sin(\widehat{\beta})\\ 
                \sin(\widehat{\beta}) & \cos(\widehat{\beta})
            \end{bmatrix*} \in SO(2). }\\
    %     &\scalemath{0.95}{
    %     \widehat{\beta}, \bm{\widehat{\xi}} = f_\phi(\bm{o}_t,\bm{o}_{t+1}), }
    % \end{split}
\end{equation}
We then learn a \gls{vo} model $f_\phi$ with parameters $\phi$ predicting $\widehat{\beta}, \bm{\widehat{\xi}}$ from observations $\bm{o}_t, \bm{o}_{t+1}$: $\widehat{\beta}, \bm{\widehat{\xi}} = f_\phi(\bm{o}_t,\bm{o}_{t+1})$
% where the \gls{vo} model $f_\phi$ is parameterized by model parameters $\phi$, processing the egocentric agent observations $\bm{o}_t$ and $\bm{o}_{t+1}$.
Finally, we transform $\widehat{\bm{g}}_t$ in coordinate system $C_{t}$ to the new agent coordinate system $C_{t+1}$ by $\widehat{\bm{g}}_{t+1}=\bm{\widehat{H}}\cdot \widehat{\bm{g}}_{t}$.

\subsection{\glsentrylong{vot}}
\glsreset{vot}
\noindent\textbf{Model Architecture: }
When facing ``optional" modalities, it is not yet clear how systems should react.
Options range from constructing an alternative input, \eg, noise~\cite{Li2022}, to falling back on a model trained without the missing modalities, to training the network with placeholder inputs~\cite{MemmelGM2021}.
Besides these, recent approaches depend on a fixed set of modalities during train- and test-time due to their \gls{cnn}-based backbone.
Transformer-based architectures can process a variable number of input tokens and can be explicitly trained to accept fewer modalities during test-time while observing multiple modalities throughout training~\cite{Vaswani2017,Bachmann2022}.
Furthermore, the Transformer's global receptive field could be beneficial for \gls{vo}, which often gets solved with correspondence or feature matching techniques~\cite{Scaramuzza2011}.
We, therefore, propose the \gls{vot}, a multi-modal Transformer-based architecture for \gls{vo}.

\noindent\textbf{Visual Odometry Estimation:}
To estimate the \gls{vo} parameters, we pass the encoded \gls{act} token to a prediction head. 
We use a two-layer \gls{mlp} with learnable parameters $\psi$ composed into $\bm{W_0}\in\mathbb{R}^{d\times d_{h}}, \bm{b}_0\in\mathbb{R}^{d_{h}}$, and $\bm{W_1}\in\mathbb{R}^{d_{h}\times 3}, \bm{b}_1\in\mathbb{R}^{3}$ with token dimensions $d=768$, and hidden dimensions $d_{h}=d/2$. A \gls{gelu}~\cite{Hendrycks2016} acts as the non-linearity between the two layers.
The \gls{vo} model can then be defined as a function $f_{\phi,\psi}(\bm{o}_t,\bm{o}_{t+1}, a_t)$ taking as input the action $a_t$ and the observations $\bm{o}_t,\bm{o}_{t+1}$ corresponding to either \rgb, \depth, or \rgbd and predicting the \gls{vo} parameters $\widehat{\beta}, \bm{\widehat{\xi}}$.
Simplifying the backbone as $b_\phi(\bm{o}_t,\bm{o}_{t+1}, a_t)$ that returns extracted visual features $\bm{v}_{t\rightarrow t+1}\in\mathbb{R}^{1\times d}$, and governed by parameters $\phi$, the resulting model is:
\begin{equation}\label{eq:vo_model_no_act}
    \begin{split}
        b_\phi(\bm{o}_t,\bm{o}_{t+1}, a_t) &= \bm{v}_{t\rightarrow t+1} \\
        \mathrm{MLP}_\psi(\bm{v}) &= \mathrm{GELU}(\bm{v}\bm{W_0} + \bm{b}_0 ) \bm{W_1} + \bm{b}_1 \\
        f_{\phi,\psi}(\bm{o}_t,\bm{o}_{t+1}, a_t) &= \mathrm{MLP}_\psi(b_\phi(\bm{o}_t,\bm{o}_{t+1},a_t)) = \widehat{\beta}, \bm{\widehat{\xi}} \\
    \end{split}
\end{equation}

\noindent\textbf{Action Prior:} The action $a_t$ taken by the agent to get from $\bm{o}_t$ to $\bm{o}_{t+1}$ is a powerful prior on the \gls{vo} parameters. To provide this information to the model, we embed the action using an embedding layer~\cite{PyTorch}. This layer acts as a learnable lookup for each action, mapping it to a fixed-size embedding.
With the embedding size equal to the token dimensions, we can create an \gls{act} and pass the information directly to the model (\cf\Cref{fig:architecture}).
In contrast to~\cite{Zhao2021,Partsey2021}, we pass the token directly to the encoder instead of concatenating it to the extracted features. This practice conditions the visual feature extraction on the action and helps ignore irrelevant parts of the image.
Note that this approach is not limited to discrete actions but tokens could represent continuous sensor readings like accelerometers, gyroscopes, and compasses, allowing for flexible deployment, \eg, in smartphones or autonomous vehicles~\cite{Srinivasa2019}.

\noindent\textbf{Explicit Modality-invariance Training:}
Explicitly training the model to be invariant to its input modalities is one way of dealing with missing sensory information during test-time.
To enforce this property, we drop modalities during training to simulate missing modalities during test-time. Furthermore, this procedure can improve training on less informative modalities by bootstrapping model performance with more informative ones. For example, \rgb is more prone to overfitting than \depth because the model can latch onto spurious image statistics, \eg textures. Training on \rgb-only would likely cause the model to latch onto those and converge to local minima, not generalizing well to unseen scenes. By increasing the amount of \depth observations seen during training, the model learns to relate both modalities, acting as regularization.
We model this notion as a multinomial distribution over modality combinations (here: \rgb, \depth, \rgbd) with equal probability. For each batch, we draw a sample from the distribution to determine on which combination to train.

\section{Experimental Evaluation}

\subsection{Setup}
\noindent\textbf{Simulation:} We use the \gls{habitat} simulator for data collection and model evaluation, following the Habitat PointNav Challenge 2020~\cite{Habitatchallenge2020sim2real} specifications.
The guidelines define an action space of \fwd (move forward $0.25m$), \turnl (turn left by $30\degree$), \turnr (turn right by $30\degree$), and \stp (indicate the agent reached its goal), and include a sensor suite of \rgbd camera, and \gps (not used in the realistic PointGoal navigation task). The \rgb observations get returned into a $[0,255]$ range while the \depth map is scaled to $[0,10]$.
Both sensors are subject to noise, \ie, noisy actuations~\cite{Pyrobot2019} and observations~\cite{Choi2015}. Furthermore, collision dynamics prevent \textit{sliding}, a behavior that allows the agent to slide along walls on collision.
Cosmetic changes bring the simulation closer to the LoCoBot~\cite{Gupta2018}, a low-cost robotic platform with an agent radius of $0.18m$ and height of $0.88m$. An optical sensor resolution of $341\times 192$ (width $\times$ height) emulates an Azure Kinect camera. An episode is successful if the agent calls \stp in a radius two times its own, \ie, $0.36m$, around the point goal and does so in $T=500$ total number of time steps.
By specification, the 3D scenes loaded into \gls{habitat} are from the Gibson~\cite{Xia2020} dataset, more precisely Gibson-4+~\cite{Savva2019}, a subset of 72 scenes with the highest quality. The validation set contains 14 scenes, which are not part of the training set.

\noindent\textbf{Dataset:} For training \gls{vot}, we collect a training- and a validation dataset. Each set consists of samples containing the ground truth translation $\bm{\xi}$ and rotation parameters $\beta$ retrieved from a perfect \gps sensor, observations $\bm{o_t}, \bm{o}_{t+1}$, and taken action $a_t$. We keep samples where the agent collides with its environment as the transformations strongly differ from standard behavior~\cite{Zhao2021}.
The collection procedure follows Zhao \etal~\cite{Zhao2021} and is performed as: 1) initialize the \gls{habitat} simulator and load a scene from the dataset, 2) place the agent at a random location within the environment with a random orientation, 3) sample a navigable PointGoal the agent should navigate to, 4) compute the shortest path and let the agent follow it, and 5) randomly sample data points along the trajectory. 
We collect \SI{250}{\kilo{}} observation-transformation pairs from the training and \SI{25}{\kilo{}} from the validation scenes of Gibson-4+, which is significantly less than comparable methods (\SI{1}{\mega{}}~\cite{Zhao2021}, \SI{5}{\mega{}}~\cite{Partsey2021}). Furthermore, we apply data augmentation during training to the \turnl and \turnr actions by horizontally flipping the observations and computing the inverse transformation.

\begin{figure*}[tb]

    \centering
    \begin{subfigure}[t]{0.32\linewidth}
        \centering
        Drop: --
    \end{subfigure}
    \begin{subfigure}[t]{0.32\linewidth}
        \centering
        Drop: \rgb
    \end{subfigure}
    \begin{subfigure}[t]{0.32\linewidth}
        \centering
        Drop: \depth
    \end{subfigure}
    \hfill
    \vfill
    \begin{subfigure}[t]{0.01\textwidth}
        \centering
        \rotatebox{90}{\small \gls{vot}-B}
    \end{subfigure}
    \hfill
    \begin{subfigure}[t]{0.31\linewidth}
        \centering
        \includegraphics[width=\textwidth]{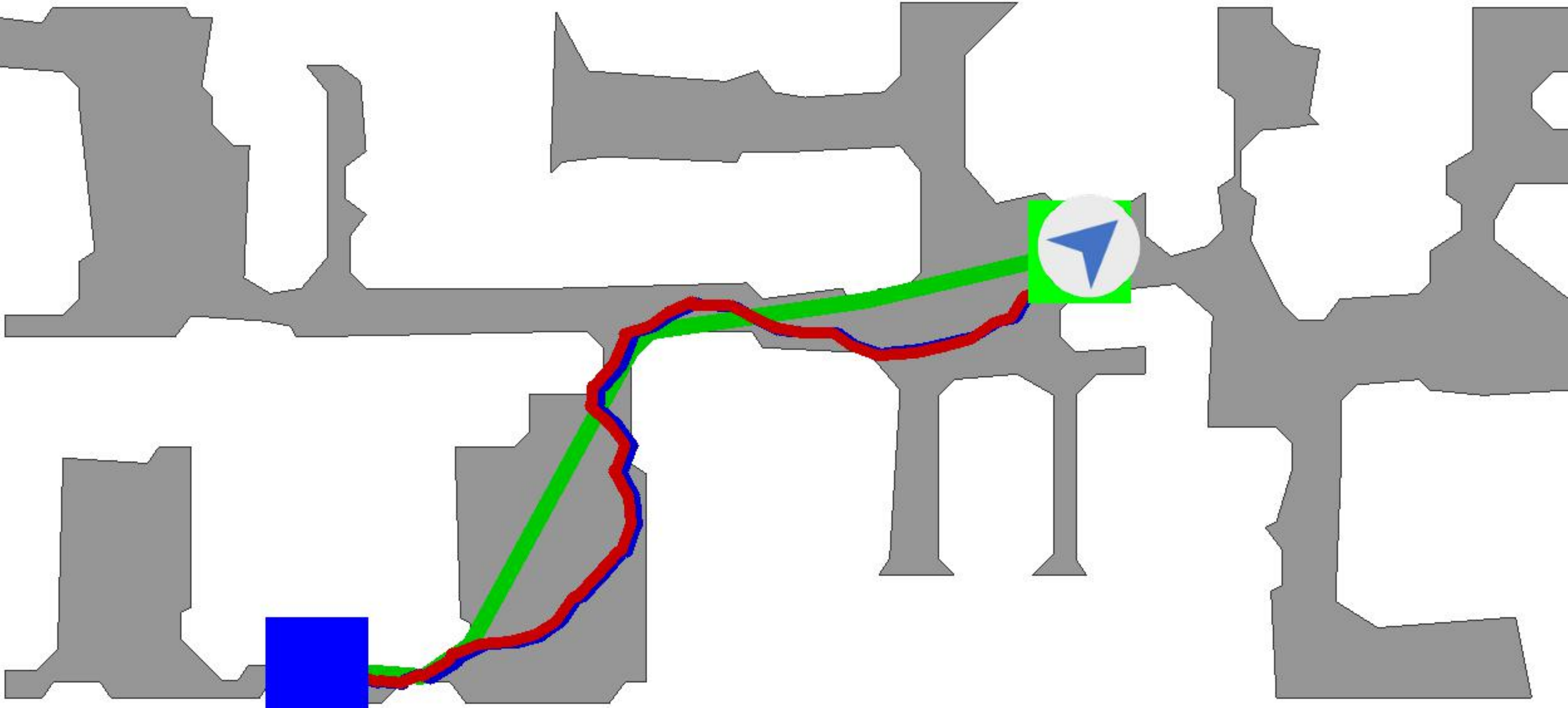}
    \end{subfigure}
    \vrule\
    \hfill
    \begin{subfigure}[t]{0.31\linewidth}
        \centering
        \includegraphics[width=\textwidth]{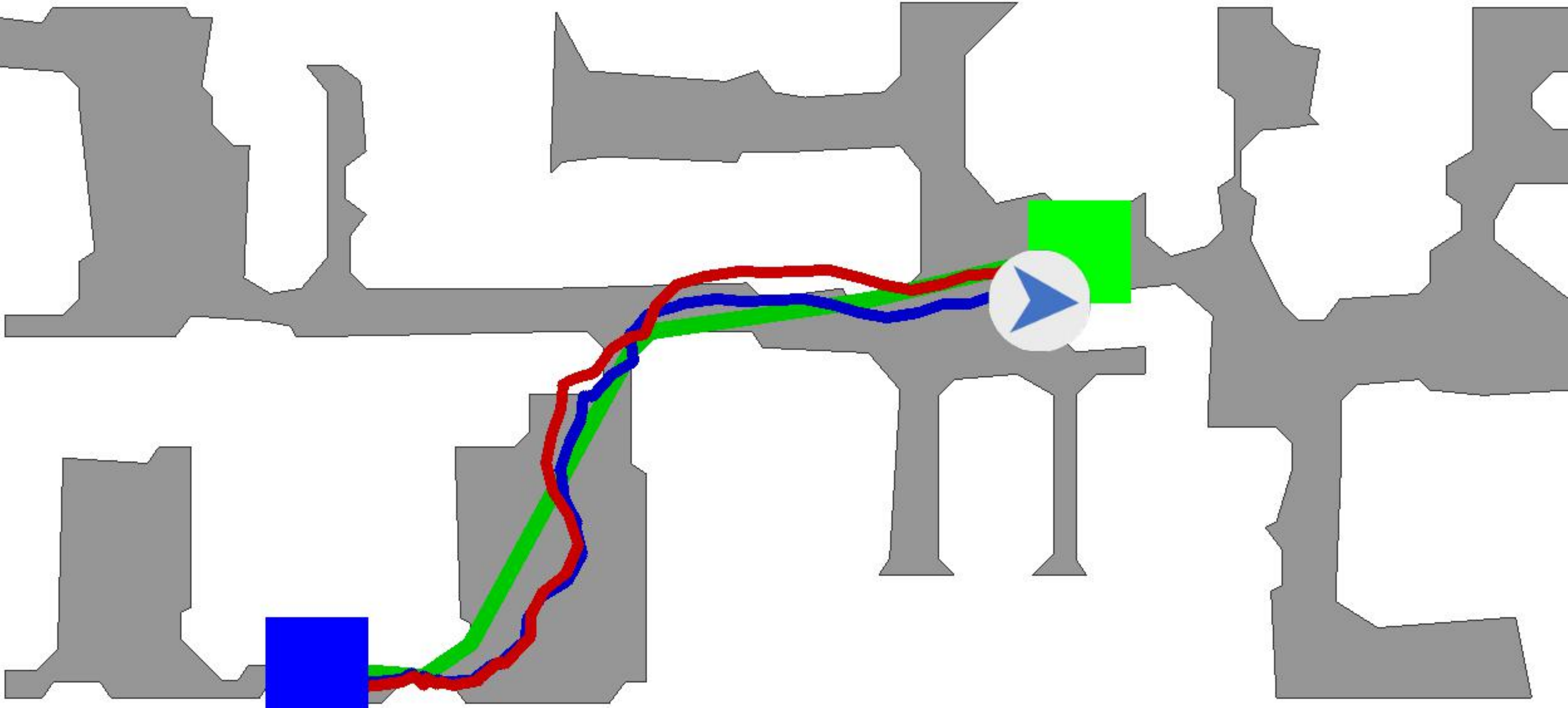}
    \end{subfigure}
    \vrule\
    \hfill
    \begin{subfigure}[t]{0.31\linewidth}
        \centering
        \includegraphics[width=\textwidth]{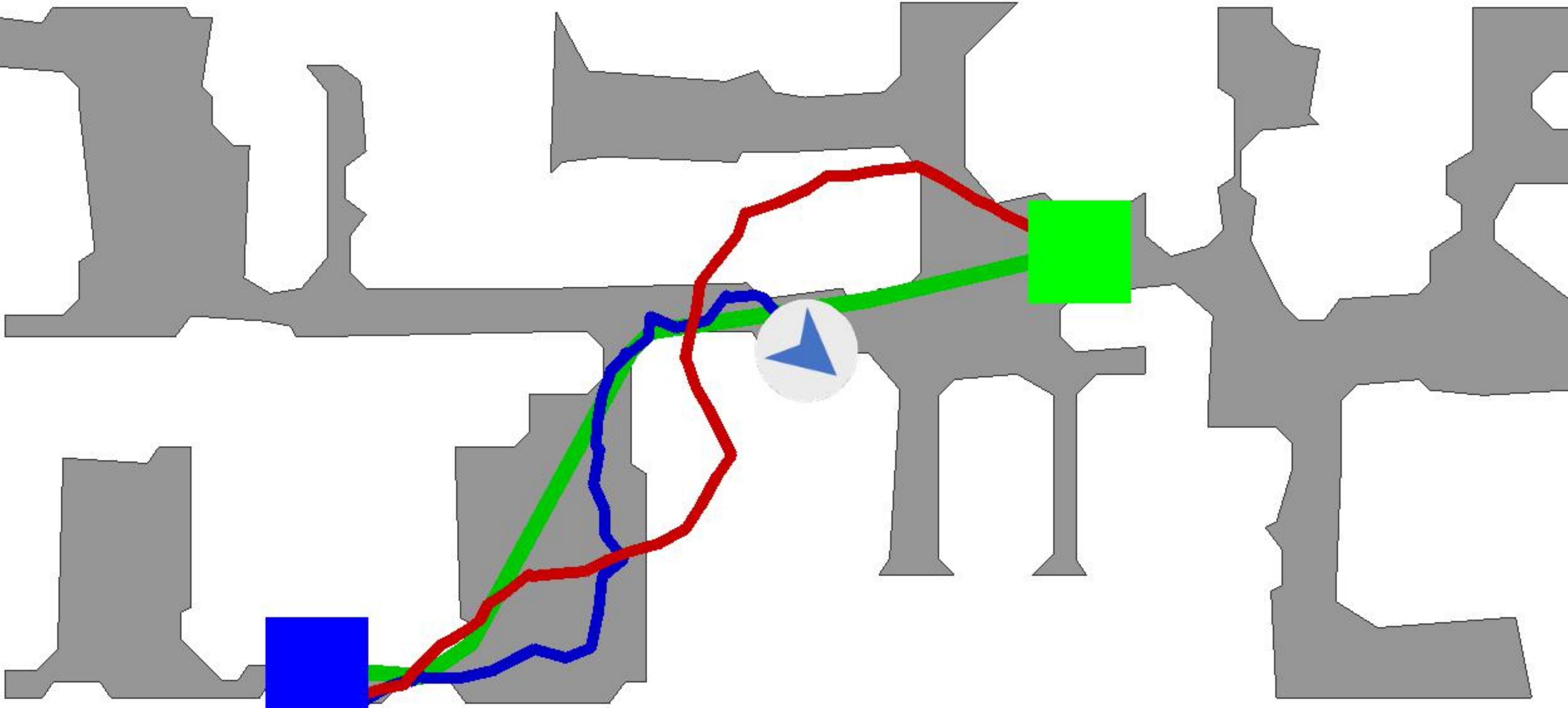}
    \end{subfigure}
    \hfill
    \vfill
    \begin{subfigure}[t]{0.01\textwidth}
        \centering
        \rotatebox{90}{\small \gls{vot}-B \textit{w/ inv.}}
    \end{subfigure}
    \hfill
    \begin{subfigure}[t]{0.31\linewidth}
        \centering
        \includegraphics[width=\textwidth]{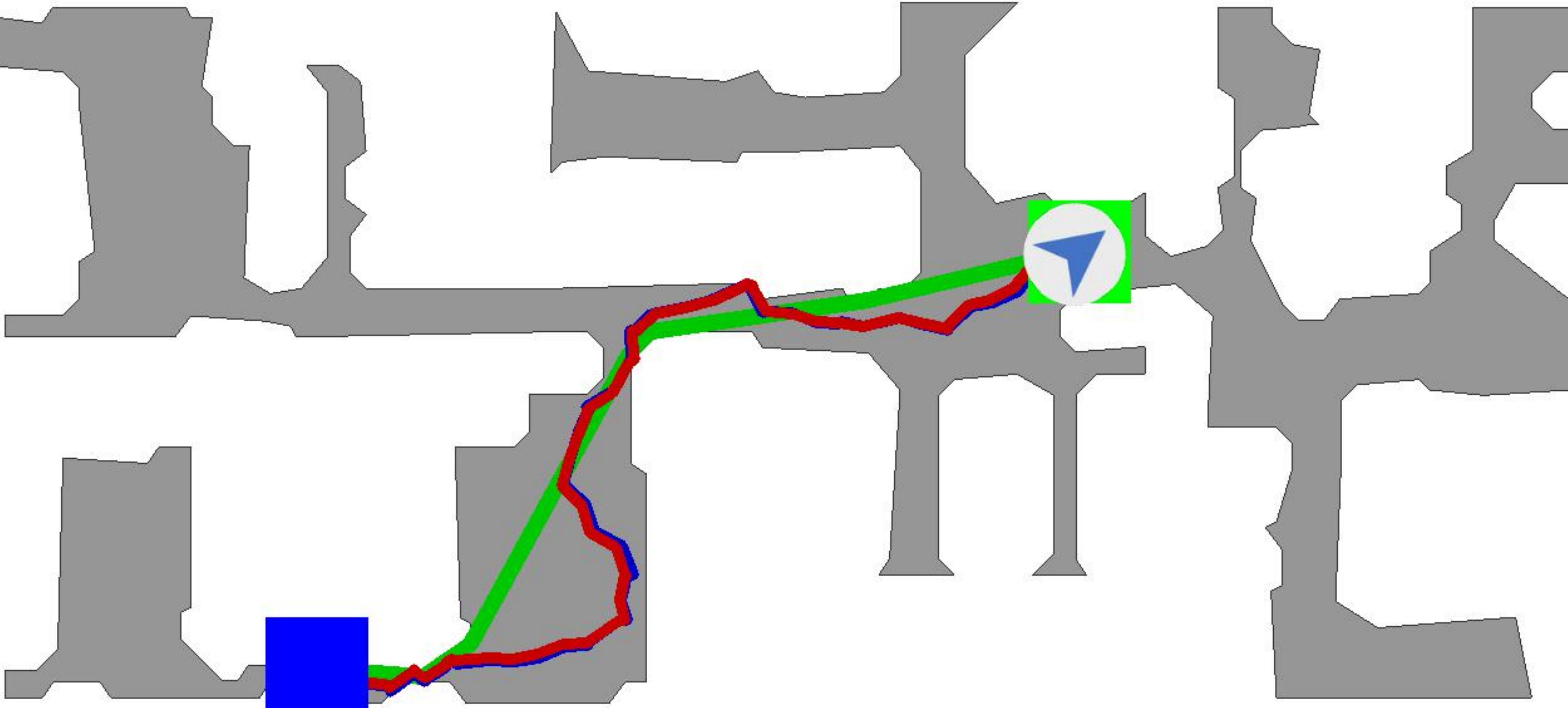}
    \end{subfigure}
    \vrule\
    \hfill
    \begin{subfigure}[t]{0.31\linewidth}
        \centering
        \includegraphics[width=\textwidth]{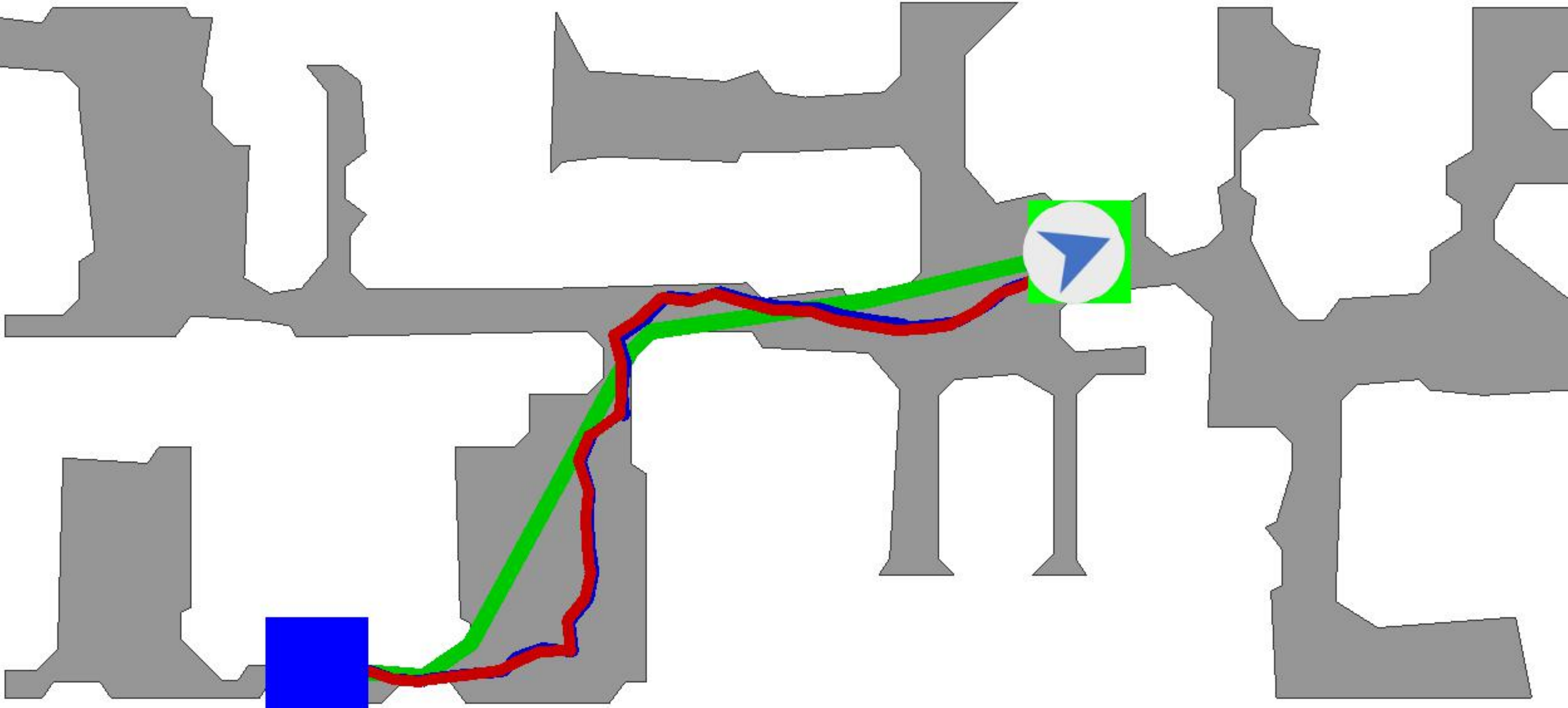}
    \end{subfigure}
    \vrule\
    \hfill
    \begin{subfigure}[t]{0.31\linewidth}
        \centering
        \includegraphics[width=\textwidth]{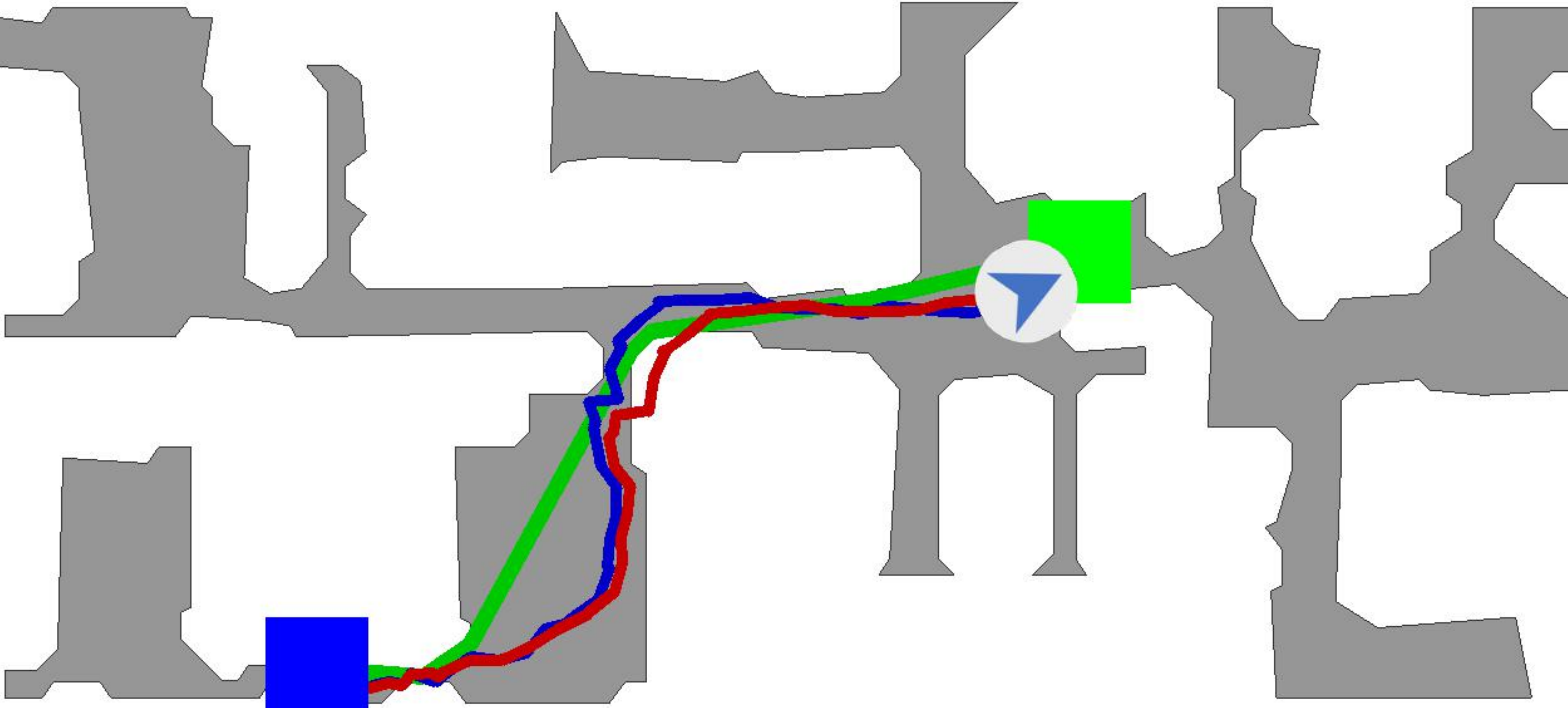}
    \end{subfigure}
    \caption{
    Top-down map of the agent navigating the \textit{Cantwell} scene~\cite{Xia2018} from start ({\color[HTML]{0001ff}\rule{0.7em}{0.7em}}) to goal ({\color[HTML]{02ff01}\rule{0.7em}{0.7em}}). The plot shows the shortest path ({\color[HTML]{3dcc5b}\rule[0.2ex]{0.7em}{0.3em}}), the path taken by the agent ({\color[HTML]{010194}\rule[0.2ex]{0.7em}{0.3em}}), and the "imaginary'' path the agent took, \ie, its \gls{vo} estimate ({\color[HTML]{ce0309}\rule[0.2ex]{0.7em}{0.3em}}). We evaluate the model without \rgb or \depth (\textit{Drop}) to determine performance when modalities are missing.
    As expected, the \gls{vot} relies heavily on both modalities, causing the estimation to drift when either \rgb or \depth is unavailable (top row). The localization error accumulates over the course of the trajectory and causes the true and imaginary path to diverge, resulting in failure to complete the episodes.
    Training a \gls{vot} to be modality-invariant (\gls{vot} \textit{w/ inv.}) removes those reliances and leads to success even when modalities are missing (bottom row).
    }
    \vspace{-1.em}
    \label{fig:navigation_paths}
    
\end{figure*}

\noindent\textbf{Loss Function:} Our loss function is the $L2$-norm between the ground truth \gls{vo} parameters and their estimated counterparts. We further add the geometric invariance losses $\mathcal{L}_{inv}$ proposed by Zhao \etal~\cite{Zhao2021} and use the Adam~\cite{Kingma2014} optimizer ($\beta_1=0.9, \beta_2=0.999, \epsilon=1e^{-8}$) to minimize the resulting loss function $\mathcal{L} = \| \bm{\xi} - \bm{\widehat{\xi}} \|^2_2 + \| \beta - \widehat{\beta} \|^2_2 + \mathcal{L}_{inv}$.

\begin{table}[tb]
    \centering
    \begin{tabular}{lllrrrr}
    \toprule
     Method & Drop & \multicolumn{1}{l}{$S\uparrow$} & \multicolumn{1}{l}{\glsentryshort{spl}$\uparrow$} & \multicolumn{1}{l}{\glsentryshort{softspl}$\uparrow$} & $d_g\downarrow$\\
    \midrule
    \glsentryshort{vot} \small \rgb & --    & 59.3 & 45.4 & 66.7 & 66.2 \\
    \glsentryshort{vot} \small \depth & --    & 93.3 & 71.7 & 72.0 & 38.0 \\
    \midrule
    \cite{Datta2020} & --    & 64.5 & 48.9 & 65.4 & 85.3 \\
    \glsentryshort{vot} & --    & 88.2 & 67.9 & 71.3 & 42.1\\
    \glsentryshort{vot} w/ \textit{inv.} & --    & \textbf{92.6} & \textbf{70.6} & \textbf{71.3} & \textbf{40.7}\\
    \midrule
    \cite{Datta2020} & \rgb & 0.0 & 0.0 & 5.4 & 398.7 \\
    \glsentryshort{vot} & \rgb & 75.9 & 58.5 & 69.9 & 59.5 \\
    \glsentryshort{vot} w/ \textit{inv.} & \rgb & \textbf{91.0} & \textbf{69.4} & \textbf{71.2} & \textbf{37.0} \\
    \midrule
    \cite{Datta2020} & \depth & 0.0 & 0.0 & 5.4 & 398.7 \\
    \glsentryshort{vot} & \depth & 26.1 & 20.0 & 58.7 & 148.1 \\
    \glsentryshort{vot} w/ \textit{inv.} & \depth & \textbf{60.9} & \textbf{47.2} & \textbf{67.7} & \textbf{72.1} \\
    \bottomrule
    \end{tabular}%
    \caption{Results for dropping modalities during test-time. Training a \gls{vot} to be modality-invariant (\textit{w/ inv.}) leads to no performance drop in comparison to a \gls{vot} trained on a single modality (\gls{vot} \rgb, \gls{vot} \depth).
    This shows that a single \gls{vot} \textit{w/ inv.} can replace multiple modality-dependent counterparts.
    Previous approaches~\cite{Datta2020,Zhao2021,Partsey2021} become inapplicable, converging to a \blind behavior. Metrics reported as $e^{-2}$. \textbf{Bold} indicates best results.}
    \label{tab:results_all_privileged}
    \vspace{-1.em}
\end{table}%

\noindent\textbf{Pre-training:}
Pre-training is a well-known practice to deal with the large data requirements of \glspl{vit}~\cite{Dosovitskiy2020,Zhai2021}, especially in a \gls{vo} setting where data is scarce~\cite{Dosovitskiy2020,Steiner2021,Kendall2015}. 
We use the pre-trained \gls{mmae} (\rgb + \depth + \gls{semseg}) made publicly available by Bachmann \etal~\cite{Bachmann2022github}. Since \gls{semseg} is unavailable in our setting, we discard the corresponding projection layers.

\noindent\textbf{Training Details:} We follow prior work~\cite{Datta2020,Zhao2021,Partsey2021} and train our navigation policy and \gls{vo} model separately before jointly evaluating them on the validation set. In contrast to \cite{Datta2020,Zhao2021}, we do not fine-tune the navigation policy on the trained \gls{vo} models as it has shown minimal navigation performance gains in~\cite{Zhao2021} and was abandoned in~\cite{Partsey2021}.

We train all models, including baselines, for 100 epochs with 10 warm-up epochs that increase the learning rate linearly from $0.0$ to $2e^{-4}$, and evaluate the checkpoints with the lowest validation error.
We further find gradient norm clipping~\cite{Zhang2019} (max gradient norm of $1.0$) to stabilize the training of \gls{vot} but to hurt the performance of the \gls{cnn} baselines.
The training was done with a batch size of 128 on an NVIDIA V100-SXM4-40GB GPU with automatic mixed-precision enabled in PyTorch~\cite{PyTorch} to reduce memory footprint and speed up training.
Our backbone is a \gls{vit}-B~\cite{Dosovitskiy2020} with a patch size of $16\times16$ and 12 encoder blocks with 12 \gls{ma} heads each, and token dimensions 768. To encode the input into tokens, we use a 2D sine-cosine positional embedding and separate linear projection layers for each modality.
Note that if additional modalities are available, our model can be extended by adding additional linear input projections or fine-tuning existing ones~\cite{Bachmann2022}.
Finally, we pass all available tokens to the model and resize each observation to $160 \times 80 \times c$ (width $\times$ height $\times$ channels c) and concatenate modalities along their height to $160 \times 160 \times c$ to reduce computation. We keep a running mean and variance to normalize \rgb and \depth to zero mean and unit variance.

\begin{figure*}[tb]
    \centering
    \begin{subfigure}[t]{0.33\textwidth}
        \centering
        \includegraphics[width=\textwidth]{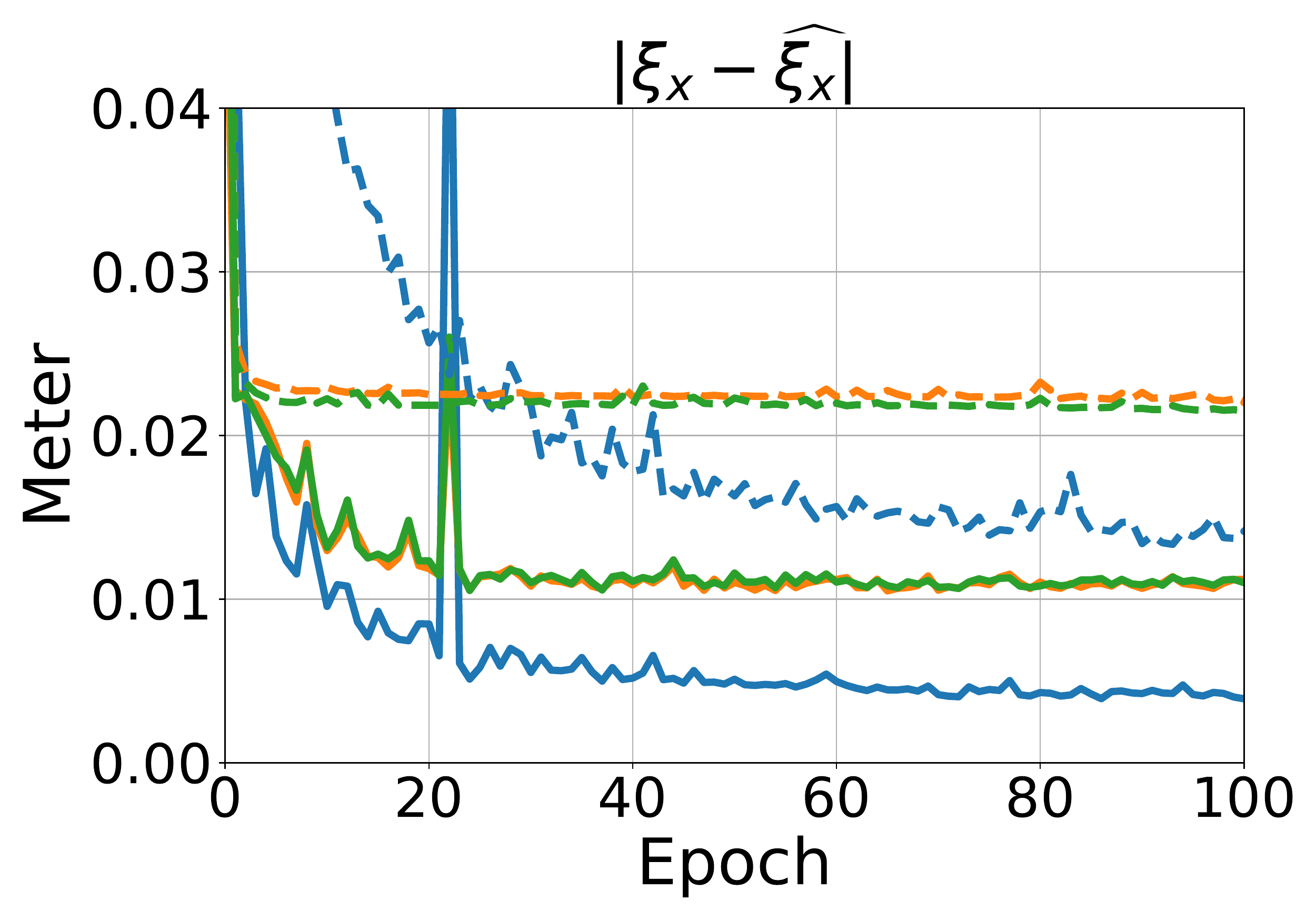}
    \end{subfigure}
    \begin{subfigure}[t]{0.33\textwidth}
        \centering
        \includegraphics[width=\textwidth]{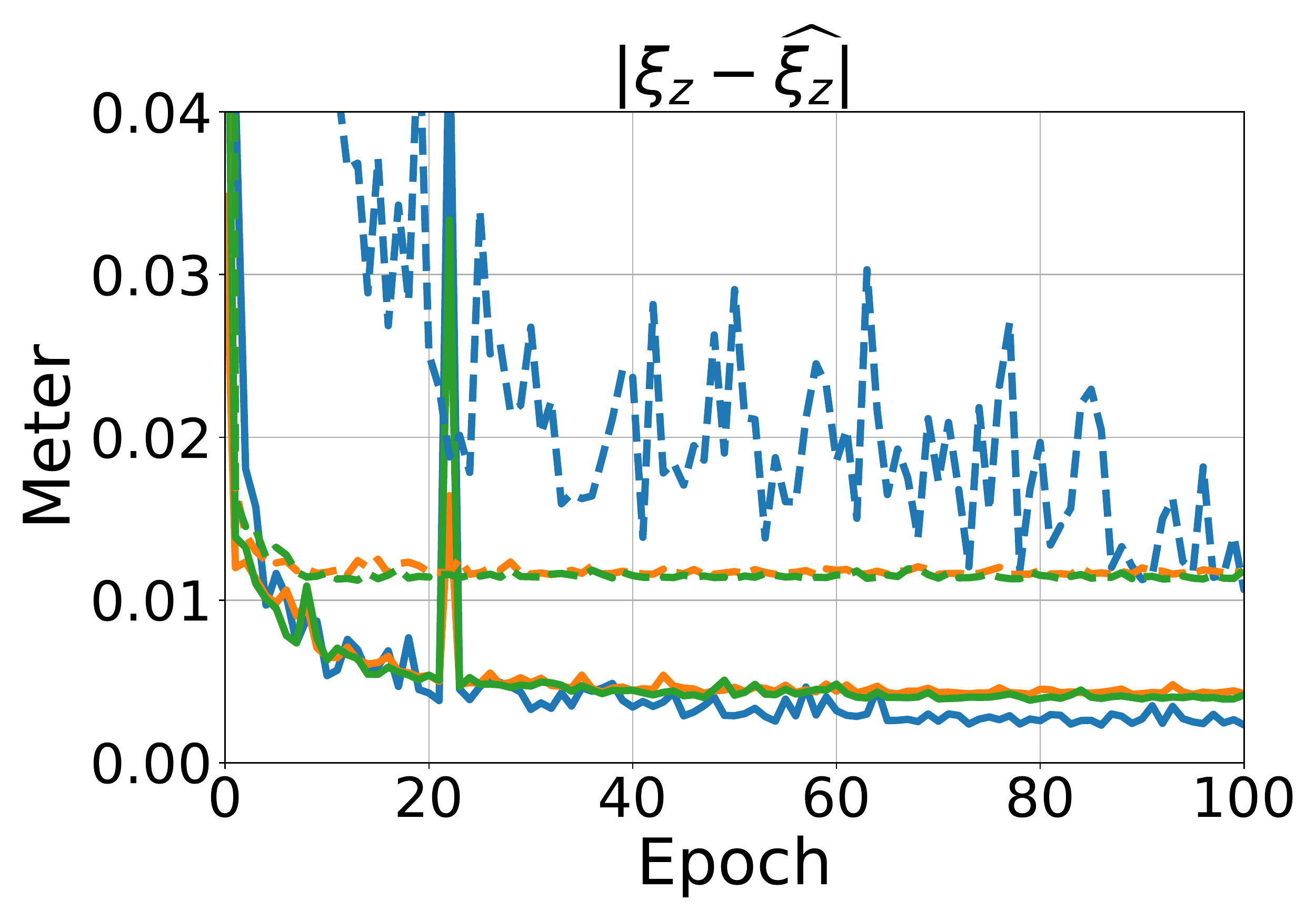}
    \end{subfigure}
    \begin{subfigure}[t]{0.33\textwidth}
        \centering
        \includegraphics[width=\textwidth]{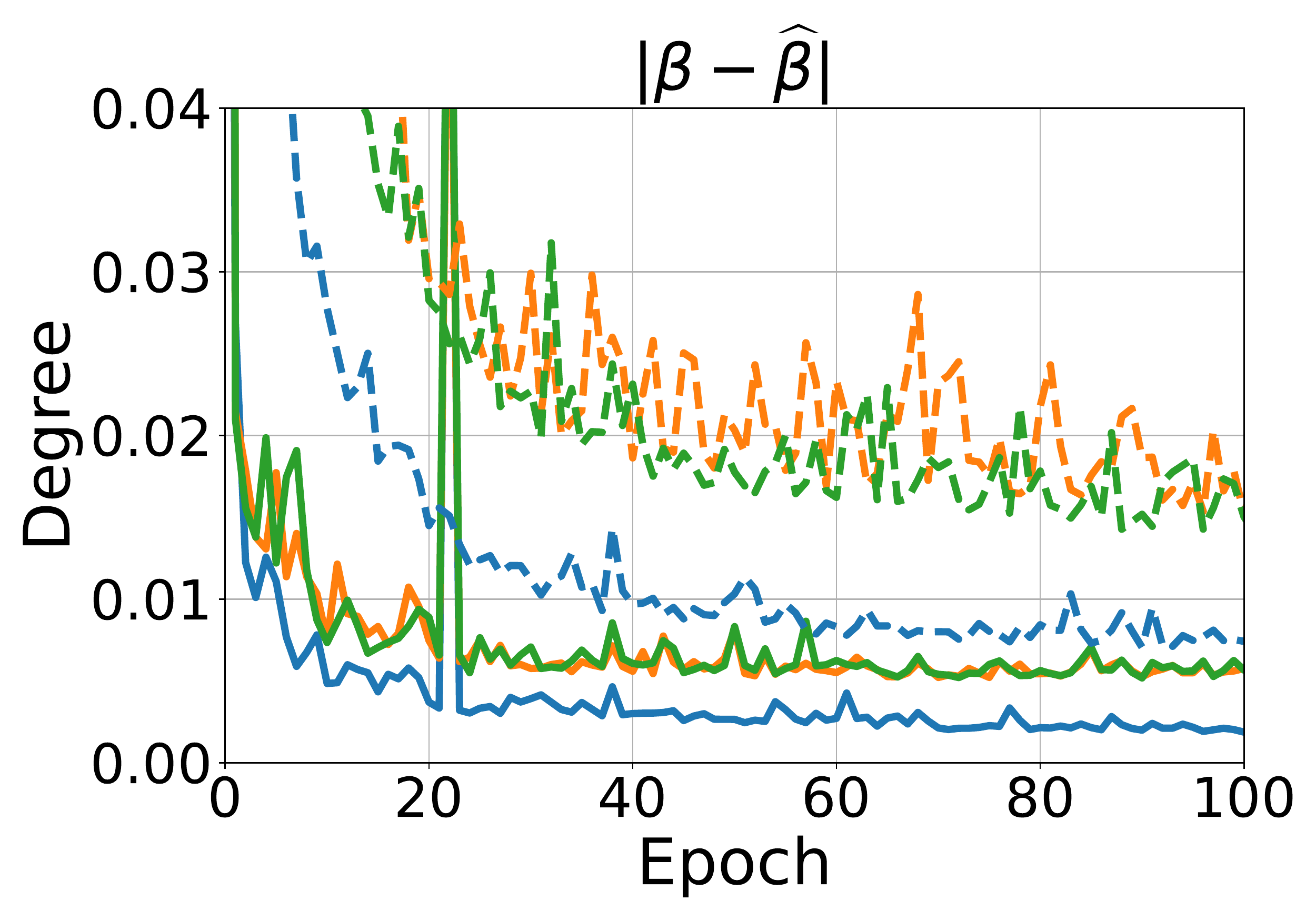}
    \end{subfigure}
    \begin{subfigure}[t]{1.\textwidth}
        \centering
        \includegraphics[width=0.6
\textwidth]{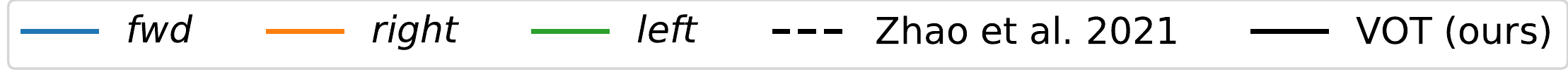}
    \end{subfigure}
    \caption{Absolute difference between ground truth translations $\bm{\xi_x}, \bm{\xi_z}$ and rotation angle $\bm{\beta}$ to their estimated counterparts $\bm{\widehat{\cdot}}$. We compare \textit{Zhao et al.}~\cite{Zhao2021} (\Cref{tab:results_all}, 2) to the \gls{vot} (\Cref{tab:results_all}, 13). Our model estimates \fwd translation along the $z$-axis (\textit{middle}), \turnl, \turnr along $z$-, $x$-axis (\textit{left, middle}), and the turning angle $\beta$ (\textit{right}) more accurately than the baseline. We successfully capture the displacements caused by the noisy actuation with an average error (over both axis $x$, $z$) of \SI{0.25}{\cm{}} (\fwd), \SI{0.7}{\cm{}} (\turnr), and \SI{0.65}{\cm{}} (\turnl).}
    \label{fig:abs_diff}
    \vspace{-1.2em}
\end{figure*}
\glsreset{spl}
\glsreset{softspl}
\noindent\textbf{Evaluation Metrics:}
Anderson \etal~\cite{Anderson2018} propose the \gls{spl} to evaluate agents in a PointGoal or ObjectGoal navigation setting.
A crucial component of this metric is the success of an episode (success $S=1$, failure $S=0$).
With $l$ the shortest path distance from the starting position and $p$ the length of the path taken by the agent, the \gls{spl} over $N$ episodes is defined as $\text{SPL}=\frac{1}{N}\sum_{i=0}^{N-1}S^{(i)}\frac{l^{(i)}}{\max(p^{(i)},l^{(i)})}$.
While \gls{spl} depends on the success of an episode, \cite{Datta2020} propose the \gls{softspl} that provides a more holistic view of the agent's navigation performance. The authors replace the binary success $S$ of an episode with a soft value consisting of the ratio between the (geodesic) distances to target upon start $d_{init}$ and termination of an episode $d_g$. The resulting metric is then $\text{SSPL} = \frac{1}{N}\sum_{i=0}^{N-1} \left(1-d_g^{(i)}/d_{init}^{(i)})\right) \frac{l^{(i)}}{\max\left(p^{(i)},l^{(i)}\right)}$
The closer the agent gets to the goal, the higher the \gls{softspl}, even if the episode is unsuccessful. This softening allows distinguishing agents that fail to complete a single or multiple episodes but move significantly close to the goal from ones that move away from it. Without access to \gps, \gls{softspl} becomes significantly more important as an agent might call \stp prematurely due to inaccurate localization.
We report the \gls{spl}, \gls{softspl}, success $S$, and (geodesic) distance to goal on termination $d_g$ on the validation scenes of Gibson-4+ with decimals truncated.

\noindent\textbf{Navigation Policy:} Similar to prior work~\cite{Datta2020,Zhao2021,Partsey2021}, we replace the \gps with our \gls{vo} model to estimate the relative goal position, which serves as the input to a pre-trained navigation policy. 
We use the same pre-trained policy as Zhao \etal~\cite{Zhao2021} for our experiments, which was trained using a goal position updated by ground truth localization.
The policy architecture consists of a \gls{lstm}\cite{Hochreiter1997} with two recurrent layers that process 1) a 512-dimensional encoding of the agent's observations $\bm{o_t}$ (here: \depth), 2) a 32-dimensional embedding of the previous action, and 3) a 32-dimensional embedding of the updated relative goal position. The observation encoding gets obtained by passing the observations $\bm{o}_t$ through a \glsentryshort{resnet}-18~\cite{He2016} backbone, flattening the resulting feature map to dimensionality 2052, and projecting it to dimensionality 512 with a fully-connected layer.
Finally, the output of the \gls{lstm} is fed through another fully-connected layer to produce a distribution over the action space and a value function estimate.
The policy was trained using DDPO~\cite{Wijmans2020}, a distributed version of \gls{ppo}~\cite{Schulman2017}.

\begin{table*}[t]

    \centering
    \resizebox{\textwidth}{!}{
    \begin{tabular}{llcccrrrrrr}
    \toprule
        & Method & Observations & Pre-train & \glsentryshort{act} & $S\uparrow$ & \glsentryshort{spl}$\uparrow$ & \glsentryshort{softspl}$\uparrow$ & $d_g\downarrow$ & $\mathcal{L}_{train}\downarrow$ & $\mathcal{L}_{val}\downarrow$ \\
        \midrule
        \midrule
        1 & \cite{Zhao2021} (separate) & \rgbd & &   & 22.4 & 13.8 & 31.5 & 305.3 & 0.125 & 0.186 \\
        2 & \cite{Zhao2021} (unified) & \rgbd & & \cmark & 64.5 & 48.9 & 65.4 & 85.3 & 0.264 & 0.420 \\
        \midrule
        \midrule
        3 & \blind & -- &   & & 0.0 & 0.0 & 5.4 & 398.7 & 48.770 & 47.258 \\
        4 & \glsentryshort{vot}-B & \rgb &   & & 27.1 & 21.2 & 57.7 & 177.0 & 0.735 & 1.075 \\ 
        5 & \glsentryshort{vot}-B & \depth &   & & {43.2} & {32.0} & {59.3} & {122.5} & \textbf{0.441} & \textbf{0.644} \\ 
        6 & \glsentryshort{vot}-B & \rgbd &   & & \textbf{47.3} & \textbf{36.3} & \textbf{61.2} & \textbf{119.7} & 1.256 & 1.698 \\ 
        \midrule
        7 & \blind & -- & & \cmark & 13.3 & 10.0 & 46.3 & 251.8 & 1.637 & 1.641 \\
        8 & \glsentryshort{vot}-B & \rgb & & \cmark & 42.0 & 32.3 & 62.7 & 107.0 & 0.043 & 0.571 \\ 
        9 & \glsentryshort{vot}-B & \depth & & \cmark & \textbf{76.1} & \textbf{58.8} & \textbf{69.2} & \textbf{60.7} & \textbf{0.017} & \textbf{0.113} \\ 
        10 & \glsentryshort{vot}-B & \rgbd & & \cmark & 72.1 & 55.6 & 68.5 & 64.4 & {0.019} & 0.129 \\
        \midrule
        \midrule
        11 & \glsentryshort{vot}-B & \rgb  & \cmark &       & 54.5  & 41.3  & 65.2  & 69.9 & 0.056 & 0.347 \\
        12 & \glsentryshort{vot}-B & \depth & \cmark &       & 83.2  & 63.4  & 69.1  & \textbf{49.9} & 0.079 & 0.205 \\
        13 & \glsentryshort{vot}-B & \rgbd & \cmark &       & \textbf{85.7}  & \textbf{65.7}  & \textbf{69.7}  & 56.1 & \textbf{0.021} & \textbf{0.060} \\
        \midrule
        14 & \glsentryshort{vot}-B & \rgb & \cmark & \cmark & 59.3 & 45.4 & 66.7 & 66.2 & 0.003 & 0.280 \\ 
        15 & \glsentryshort{vot}-B & \depth & \cmark & \cmark & \textbf{93.3} & \textbf{71.7} & \textbf{72.0} & \textbf{38.0} & \textbf{0.004} & \textbf{0.044} \\
        16 & \glsentryshort{vot}-B & \rgbd & \cmark & \cmark & 88.2 & 67.9 & 71.3 & 42.1 & 0.004 & 0.051 \\ 
        \midrule
        \midrule
        17 & \glsentryshort{vot}-B \textit{w/ inv.} & \rgbd & \cmark & \cmark & \textbf{92.6} & \textbf{70.6} & \textbf{71.3} & \textbf{40.7} & \textbf{0.008} & \textbf{0.094} \\
        \midrule
        \midrule
        & \textit{oracle} & \gps & -- & -- & {97.8} & {74.8} & {73.1} & {29.9} & -- & -- \\
    \bottomrule
    \end{tabular}
    }
    \caption{Ablation study of architecture design and input modalities. We further investigate pre-training with \gls{mmae}~\cite{Bachmann2022} in models 11-14. Losses $\mathcal{L}$, Success $S$, \glsentryshort{spl}, \glsentryshort{softspl}, and $d_g$ reported as $e^{-2}$. \textbf{Bold} indicates best results.}
    \label{tab:results_all}
    \vspace{-1.em}
\end{table*}

\subsection{Dealing With Optional Modalities}
We evaluate the models' robustness to missing modalities by randomly dropping access to one of the training modalities.
%We evaluate the models' reaction to ``optional" modalities by randomly dropping access to one of the training modalities.
This setup probes \gls{vot} for dependencies on the input modalities, which directly influence the downstream performance under limited access.
In case of sensor malfunctioning, \eg, only a single modality might be available, a \gls{cnn}'s failure is predetermined as it requires a fixed-size input. If not given, the system converges to a \blind behavior, exemplified in~\Cref{tab:results_all_privileged}.
Limiting access to modalities reveals \gls{vot}'s dependency on \depth. Dropping \rgb barely decreases performance, while dropping \depth causes the localization to fail more drastically. Comparing the true agent localization and its "imaginary'', \ie, \gls{vo} estimate, it becomes clear why. \Cref{fig:navigation_paths} shows how the errors accumulate, causing the true location to drift away from the estimate.
While the effect is less drastic when dropping \rgb, the agent still fails to reach the goal.

Training \gls{vot} with the proposed invariance training (\textit{w/ inv.}), \ie, sampling \rgb for 20\%, \depth for 30\%, and \rgbd for 50\% of the training batches, eliminates this shortcoming.
Removing \rgb now only decreases the success rate by $1.6\%$, while removing \depth also leads to a stronger performance. This observation suggests that \rgb is less informative for the \gls{vo} task than \depth.
Especially when navigating narrow passages, \rgb might consist of uniform observations, \eg, textureless surfaces like walls, making it hard to infer the displacement, unlike \depth which would still provide sufficient geometric information (\cf \Cref{fig:navigation_paths}).
However, this information asymmetry only leads to a decline in the metrics that are sensitive to subtle inconsistencies in the localization, \ie, $S$, and \gls{spl}. Inspecting the \gls{softspl}, the drop of $-3.5$ is less drastic.
Explicit modality-invariance training keeps \glsentryshort{vot}-B~(\rgbd) from exploiting this asymmetry and matches the performance of \glsentryshort{vot}-B~(\rgb) when \depth is dropped during test-time~\cref{tab:results_all_privileged}.

\begin{table}[t]

    \centering
    \resizebox{\columnwidth}{!}{
    \begin{tabular}{cccccc}
    \toprule
    Rank &  Participant team &  S & \gls{spl} & \gls{softspl} & $d_g$ \\
    \midrule
    1 & \textbf{MultiModalVO (VOT)} (ours) & 93 & 74 & 77 & 21 \\ 
    2 & VO for Realistic PointGoal~\cite{Partsey2021} & 94 & 74 & 76 & 21 \\ 
    3 & inspir.ai robotics & 91 & 70 & 71 & 70 \\ 
    4 & VO2021~\cite{Zhao2021} & 78 & 59 & 69 & 53 \\ 
    5 & Differentiable SLAM-net~\cite{Karkus2021} & 65 & 47 & 60 & 174 \\ 
    \bottomrule
    \end{tabular}%
    }
    \caption{\textbf{Habitat Challenge 2021.} Results for the Point Nav Test-Standard Phase (test-std split) retrieved on 05-Nov-2022.}
    \label{tab:results_habitat_challenge}
    \vspace{-1.5em}
\end{table}%
% \input{images/attention_maps.tex}

% \subsection{Baseline Comparison}
\subsection{Quantitative Results}
We compare our approach to Zhao \etal~\cite{Zhao2021} in terms of downstream navigation performance, \ie, the \gls{vo} model as \gps replacement for a learned navigation agent. We use the same publicly available navigation policy for both approaches and the published \gls{vo} models of the baseline~\cite{Zhao2021}.
Using only 25\% of the training data, \gls{vot} improves performance by $S+12.3$, \gls{spl}$+9.7$, \gls{softspl}$+2.0$ (\cf \Cref{tab:results_all} 15) and $S+7.2$, \gls{spl}$+5.7$, \gls{softspl}$+1.3$ (\cf \Cref{tab:results_all} 16).
When training the baseline on our smaller data set (\cf \Cref{tab:results_all} 2, unified, \glsentryshort{resnet}-50), this improvement increases to $S+29.8$, \gls{spl}$+22.8$, \gls{softspl}$+6.6$ (\cf \Cref{tab:results_all} 15) and $S+23.7$, \gls{spl}$+19.0$, \gls{softspl}$+5.9$ (\cf \Cref{tab:results_all} 16).

To capture the raw \gls{vo} performance detached from the indoor navigation task, we inspect the absolute prediction error in \Cref{fig:abs_diff}. We differentiate between translation $\bm{\xi}$ in $x$- and $y$- direction ($\xi_x$, $\xi_y$), and taken action.
\gls{vot} is accurate up to \SI{0.36}{\cm{}} (\fwd), \SI{1.04}{\cm{}} (\turnr), \SI{1.05}{\cm{}} (\turnl) in $x$- direction and \SI{0.20}{\cm{}} (\fwd), \SI{0.41}{\cm{}} (\turnr), \SI{0.38}{\cm{}} (\turnl) in $z$-direction. Note how the baseline struggles to capture $\xi_z$, corresponding to the forward-moving direction $z$ when taking the \fwd action.

Given the results in \Cref{tab:results_all}, we advise using \gls{vot} trained on \depth-only when access is assumed, as the difference to using \gps is a mere $S-4.5$, \gls{spl}$-3.1$, \gls{softspl}$-1.1$. When "optional'' modalities are needed, \eg, they are expected to change during test-time, invariance training should be used. Trained on \rgbd, this setup also reaches \gps like performance with differences of only $S-5.2$, \gls{spl}$-4.2$, \gls{softspl}$-1.8$.

\begin{figure*}[tb]

\begin{subfigure}[t]{0.5\textwidth}
    \centering
    \begin{subfigure}[t]{0.24\linewidth}
        \centering
        \small $o_t$ (\rgb)
    \end{subfigure}
    \begin{subfigure}[t]{0.24\linewidth}
        \centering
        \small $o_{t+1}$ (\rgb)
    \end{subfigure}
    \begin{subfigure}[t]{0.24\linewidth}
        \centering
        \small $o_t$ (\depth)
    \end{subfigure}
    \begin{subfigure}[t]{0.24\linewidth}
        \centering
        \small $o_{t+1}$ (\depth)
    \end{subfigure}
    \hfill
    \vfill
    \begin{subfigure}[t]{0.24\linewidth}
        \centering
        \includegraphics[width=\textwidth]{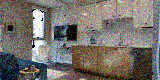}
    \end{subfigure}
    \begin{subfigure}[t]{0.24\linewidth}
        \centering
        \includegraphics[width=\textwidth]{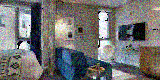}
    \end{subfigure}
    \begin{subfigure}[t]{0.24\linewidth}
        \centering
        \includegraphics[width=\textwidth]{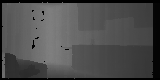}
    \end{subfigure}
    \begin{subfigure}[t]{0.24\linewidth}
        \centering
        \includegraphics[width=\textwidth]{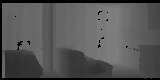}
    \end{subfigure}
    \hfill
    \vfill
    \begin{subfigure}[t]{0.24\linewidth}
        \centering
        \includegraphics[width=\textwidth]{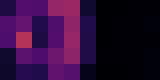}
    \end{subfigure}
    \begin{subfigure}[t]{0.24\linewidth}
        \centering
        \includegraphics[width=\textwidth]{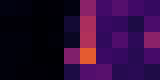}
    \end{subfigure}
    \begin{subfigure}[t]{0.24\linewidth}
        \centering
        \includegraphics[width=\textwidth]{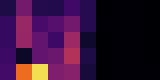}
    \end{subfigure}
    \begin{subfigure}[t]{0.24\linewidth}
        \centering
        \includegraphics[width=\textwidth]{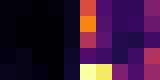}
    \end{subfigure}
    \hfill
    \vfill
    \caption{Ground truth action: \turnl}
    \label{fig:attention_vit_b_left}
\end{subfigure}
\begin{subfigure}[t]{0.5\textwidth}
    \centering
    \begin{subfigure}[t]{0.24\linewidth}
        \centering
        \small $o_t$ (\rgb)
    \end{subfigure}
    \begin{subfigure}[t]{0.24\linewidth}
        \centering
        \small $o_{t+1}$ (\rgb)
    \end{subfigure}
    \begin{subfigure}[t]{0.24\linewidth}
        \centering
        \small $o_t$ (\depth)
    \end{subfigure}
    \begin{subfigure}[t]{0.24\linewidth}
        \centering
        \small $o_{t+1}$ (\depth)
    \end{subfigure}
    \hfill
    \vfill
    \begin{subfigure}[t]{0.24\linewidth}
        \centering
        \includegraphics[width=\textwidth]{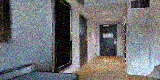}
    \end{subfigure}
    \begin{subfigure}[t]{0.24\linewidth}
        \centering
        \includegraphics[width=\textwidth]{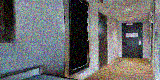}
    \end{subfigure}
    \begin{subfigure}[t]{0.24\linewidth}
        \centering
        \includegraphics[width=\textwidth]{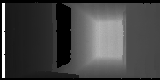}
    \end{subfigure}
    \begin{subfigure}[t]{0.24\linewidth}
        \centering
        \includegraphics[width=\textwidth]{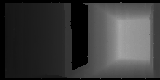}
    \end{subfigure}
    \hfill
    \vfill
    \begin{subfigure}[t]{0.24\linewidth}
        \centering
        \includegraphics[width=\textwidth]{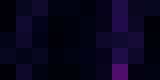}
    \end{subfigure}
    \begin{subfigure}[t]{0.24\linewidth}
        \centering
        \includegraphics[width=\textwidth]{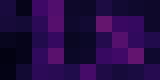}
    \end{subfigure}
    \begin{subfigure}[t]{0.24\linewidth}
        \centering
        \includegraphics[width=\textwidth]{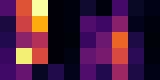}
    \end{subfigure}
    \begin{subfigure}[t]{0.24\linewidth}
        \centering
        \includegraphics[width=\textwidth]{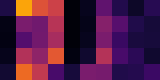}
    \end{subfigure}
    \hfill
    \vfill
    \caption{Ground truth action: \fwd}
    \label{fig:attention_vit_b_fwd}
\end{subfigure}

\caption{
    Attention maps of the last attention layer of \gls{vot} (\cf \Cref{tab:results_all} 13).
    Brighter color indicates higher ({\color[HTML]{f1ee74}\rule{0.7em}{0.7em}}) and darker color lower ({\color[HTML]{010100}\rule{0.7em}{0.7em}}) weighting of the image patch. The \gls{vot} learns to focus on regions present in both time steps $t,t+1$, \ie, outer image regions for turning \turnl, and center regions for moving \fwd. Artifacts of the Gibson dataset get ignored (\cf\Cref{fig:attention_vit_b_fwd}).
}
\label{fig:attention_vit_b}
\vspace{-1.5em}
\end{figure*}

\subsection{Ablation Study}

We identify the impact of different input modalities and model design choices in our ablation study (\cf~\Cref{tab:results_all}).
Without observations, the \blind \gls{vo} model cannot update the goal position. This means the agent can only act without goal-related feedback, resulting in a $0\%$ success rate.

Extending the model with our proposed \gls{act} token allows it to surpass the \blind performance. Able to update the relative goal position, the agent reaches an \gls{softspl} of $46.3$, but due to the actuation noise, it calls \stp correctly only $13.3\%$ of the time.
Access to \rgb or \depth allows the \gls{vo} model to adjust to those unpredictable displacements. While the \rgb and \depth observations correlate with the \gls{act} token, they also contain information about the noisy actuation.
Vice versa, \gls{act} disambiguates corner cases where the visual observations do not provide explicit information about the underlying action. For instance, a \fwd action colliding with a wall might be hard to distinguish from a noisy \turnl turning less than $30\degree$~\cite{Zhao2021}.

Our results show that \gls{mmae} pre-training provides useful multi-modal features for \gls{vo} that fine-tuned outperform the \gls{cnn} baselines. In addition, these features are complementary to the \gls{act} prior, together achieving state-of-the-art results. We conclude that the \gls{act} prior biases the model towards the mean of the corresponding transformation, while the pre-training supports the learning of the additive actuation noise.

Training separate models for each modality reveals that \depth is a more informative modality than \rgb for \gls{vo}. We assume this to be a direct result of its geometric properties, \ie, the 3D structure of the scene. We find that training \gls{vot} on noisy \rgb even hurts the localization. The model overfits the visual appearance of the scenes and is unable to generalize to unseen ones. In turn, \depth does not suffer from this issue as it only contains geometric information.

\subsection{Action-conditioned Feature Extraction}
We show what image regions the model attends to by visualizing the attention maps of the last \gls{ma}-layer (\cf \Cref{tab:results_all} 16) corresponding to the \gls{act} token in \Cref{fig:attention_vit_b}.
To reduce the dimensionality of the visualization, we fuse the heads' weights via the $max$ operator and align the attention maps with the input images. We normalize the maps to show the full range of the color scheme.

We find that passing different actions to \gls{vot} primes it to attend to meaningful regions in the image.
When passed turning actions \turnl or \turnr, \gls{vot} focuses on regions present at both time steps.
This makes intuitive sense, as a turning action of $30\degree$ strongly displaces visual features or even pushes them out of the agent's field of view.
A similar behavior emerges for a \fwd action which leads to more attention on the center regions, \eg, the walls and the end of a hallway (\cf~\Cref{fig:attention_vit_b_fwd}).
These results are particularly interesting as the model has no prior knowledge about the \gls{vo} task but learns something about its underlying structure.

\subsection{Habitat Challenge 2021 PointNav}
We compare our approach (\cf \Cref{tab:results_all} 16) to several baselines submitted to the \textit{Habitat Challenge 2021} benchmark in \Cref{tab:results_habitat_challenge}.
Using the same navigation policy as Partsey \etal~\cite{Partsey2021}, \gls{vot} achieves the highest \gls{softspl} and on par \gls{spl} and $d_g$ training on only 5\% of the data. These results clearly show that reusability doesn't come with a price of lower performance and that scaling data requirements doesn't seem to be the answer to solving deep \gls{vo}.

\subsection{Limitations}
In our work, we separate the \gls{vo} model from the navigation policy and only focus on the modality-invariance of the former, neglecting that the navigation policy expects \depth as input~\cite{Datta2020,Zhao2021,Partsey2021}. Designing policies to be modality-invariant is subject to future research.
Assuming an accurate sensor failure detection when dropping modalities, additionally, is an idealized setup.
Furthermore, our experiments in the \gls{habitat}'s simulator limit the available modalities to \rgbd. Even though \gls{semseg} has shown to be beneficial for some \gls{vo} applications~\cite{Valada2018,Radwan2018}, there is no specific sensor for it. However, \gls{semseg} could be estimated from \rgb.
While our experiments focus on discrete actions and \rgbd, our architecture could be adapted to continuous actions and other sensor types. However, training might become more difficult due to a lack of pre-trained weights.
% Finally, \gls{vot} is only as good as the input modalities during test-time, \ie, training the model to be modality invariant cannot extract more information from a modality than a model trained on that one alone.

\section{Conclusions}
We present \glsentrylongpl{vot} for learned \glsentrylong{vo}. Through multi-modal pre-training and action-conditioned feature extraction, our method is sample efficient and outperforms current methods trained on an order of magnitude more data.
With its modality-agnostic design and modality-invariance training, a single model can deal with different sensor suites during training and can trade-off subsets of those during test-time.
% Our contributions pave the way for reusable and widely applicable \gls{vo} models.

% \newpage
%%%%%%%%% REFERENCES
{\small
\bibliographystyle{ieee_fullname}
\bibliography{bibliography}
}

\end{document}